\documentclass{article}
\usepackage[table]{xcolor}
\usepackage{graphicx}
\usepackage{stackengine}
\usepackage{enumitem}

\def\bx{\mathbf{x}}

\def\bb{\mathbf{b}}

\def\bg{\mathbf{g}}
\def\bX{\mathbf{X}}

\def\bW{\mathbf{W}}

\newcommand{\be}{\begin{equation}}
\newcommand{\ee}{\end{equation}}
\newcommand{\bel}{\begin{equation}}
\newcommand{\eel}{\end{equation}}
\newcommand{\bea}{\begin{eqnarray}}
\newcommand{\eea}{\end{eqnarray}}
\newcommand{\beal}{\begin{eqnarray}}
\newcommand{\eeal}{\end{eqnarray}}

\usepackage{amsmath}
\usepackage{amssymb}
\usepackage{algorithm,algorithmic}
\usepackage{mathtools}
\usepackage{multirow}

\usepackage{microtype}
\usepackage{graphicx}
\usepackage{subfigure}
\usepackage{booktabs} 

\usepackage{hyperref}
\usepackage{xr} 
\usepackage[accepted]{icml2019}
\usepackage{subfigure}

\icmltitlerunning{Feature Grouping as a Stochastic Regularizer for High-Dimensional Structured Data}

\begin{document}

\twocolumn[
\icmltitle{Feature Grouping as a Stochastic Regularizer \\ for High-Dimensional Structured Data}




\begin{icmlauthorlist}
\icmlauthor{Serg{\"u}l Ayd{\"o}re}{stevens}
\icmlauthor{Bertrand Thirion}{parietal}
\icmlauthor{Ga{\"e}l Varoquaux}{parietal}
\end{icmlauthorlist}

\icmlaffiliation{stevens}{Stevens Institute of Technology, New Jersey,
USA}
\icmlaffiliation{parietal}{Inria Saclay, Palaiseau, France}

\icmlcorrespondingauthor{Sergul Aydore}{saydore@stevens.edu}

\icmlkeywords{regularization, high-dimensional data}

\vskip 0.3in
]




\printAffiliationsAndNotice{}  

\begin{abstract}
In many applications where collecting data is expensive, for example
neuroscience or medical imaging, the sample size is typically small compared
to the feature dimension.
It is challenging in this setting to train expressive, non-linear models without overfitting.
These datasets call for intelligent regularization that
exploits known structure, such as correlations
between the features arising from the measurement device.
However, existing structured regularizers need specially crafted
solvers, which  are difficult to apply to complex models.
We propose a new regularizer specifically designed to leverage
structure in the data in a way that can be applied efficiently to complex
models.
Our approach relies on feature grouping, using a fast clustering algorithm inside a stochastic gradient descent loop:
given a family of feature groupings that capture feature covariations, we randomly select these groups at each iteration.
We show that this approach amounts to enforcing a denoising regularizer on the solution.
The method is easy to implement in many model architectures, such as
fully connected neural networks, and has a linear computational cost.
We apply this regularizer to a real-world fMRI dataset and the Olivetti Faces datasets.
Experiments on both datasets demonstrate that the proposed approach produces
 models that generalize better than those trained with conventional regularizers, and also
improves convergence speed.
\end{abstract}

\section{Introduction}
\label{Introduction}

Fitting complex machine learning (ML) models has lead
to impressive gains in accuracy in various fields, such as computer
vision, speech processing, and natural language processing
\cite{lecun2015deep, mnih2013playing}.
Yet, the success of complex models has not carried over to high-dimensional
small-sample data such as full-brain images, despite clear potential
\cite{plis2014deep, suk2016state}.
Indeed, complex models are prone to \textit{overfitting} in settings such as
those encountered in neuroimaging:
\begin{enumerate}[leftmargin=3ex, itemsep=.2ex, parsep=.1ex, topsep=.1ex]
\item \underline{Large feature dimension:} Neuroimaging data are
very high-dimensional, due to the progress in image resolution.
For example, functional Magnetic Resonance Images (fMRIs) \footnote{fMRI is a noninvasive
neuroimaging modality that measures brain activity during cognitive tasks
in humans.} are represented by 4D-arrays of 3D images over time.
The total dimensionality is in the order of $10^7$. This leads to the 
phenomenon known as the \textit{curse of dimensionality}, and is often an 
obstacle so the success of ML. 
\noindent
\item \underline{Noise in the data:} Neuroimaging data contain a
significant amount of physiological, respiratory, and mechanical
artifacts unrelated to the effect of interest. Removal of this noise is a
difficult task. Ideally, we need an ML model that is robust against noise.
\noindent
\item \underline{Small sample size:} Neuroimaging data typically
have small sample sizes due to the logistics and cost of data acquisition,
as well as the effort
required to recruit subjects. It takes several hours to collect
data from a single individual. Therefore, the number of examples in 
neuroimaging data
is usually in the order of hundreds, as opposed to other ML applications,
such as computer vision, in which modern data sets comprise at least
hundreds of thousands samples.
\end{enumerate}
These challenges are not limited to neuroimaging
applications. They are common in medical imaging, genomics,
chemistry, and financial applications \cite{fan2006statistical, 10002015global}.
\textit{Regularization} is crucial for the success of ML in such settings.
The optimal regularization strategy for a given dataset should
leverage the known structure of the data. Yet, classic approaches to
structured regularization \cite{bach2012structured} entail high computational 
cost and are not-well suited for fitting complex models with stochastic
gradient descent.
Here, we introduce a \textit{structured regularization} strategy
integrated in a \textit{stochastic gradient descent} (SGD) loop to tackle the challenges described above.

\subsection{Related works: strategies to tackle overfit}

A long-standing body of work tackle overfit in ML models. Here, we briefly
mention approaches that are most closely related to the present effort.
Conventional approaches to mitigate overfitting include penalizing model
weights, seeking a reduced-dimensionality parametrization, or sharing
weights between related inputs or outputs.
%
%
%

Regularization with $\ell_1$ or $\ell_2$ penalties
\cite{tibshirani1996regression} reduces overfitting by biasing
weights to avoid large values due to chance.
In  high-dimensional data, features often display 
groups of highly correlated or irrelevant features
\cite{buhlmann2013correlated}.
Structured penalties \cite{bach2012structured} leverage 
a priori hypotheses on these groups, fostering sparsity accordingly.
These approaches are based on the group lasso
\cite{yuan2006model} which generalizes $\ell_1$ regularization to groups
of features. \citet{zhao2009composite} 
similarly generalize $\ell_2$ regularization to capture
feature groups. 
A drawback of these formulation is that they require groups of features
to be manually identified. Using the overlapping group lasso \cite{jacob2009group}
 enables more systematic definitions of groups \cite{bach2012structured}.
However, as the dimension grows, the number of overlapping groups gives
rise to prohibitive computational costs. In addition, these approaches
are limited to convex models.

Feature grouping by actually merging the features into a single variable
gives faster algorithms, though these are not formulated as a single
optimization and rely on heuristics \cite{garcia2016high}. 
Using a clustering algorithm to group features is a long-standing
dimensionality reduction technique used for ML on high-dimensional data
\cite{mccallum2000efficient, thalamuthu2006evaluation, xu2005survey}. 
Combining it with model ensembling gives more robustness to the feature
grouping \cite{varoquaux2012small}.

In general, a good dimensionality reduction can limit overfit and
improve prediction of a model by reducing its input dimensionality, and
thus the number of model parameters. Using random matrices to project
data onto a lower dimensional space can capture the important properties of the data \cite{bingham2001random,
achlioptas2003database} and
thus give very computationally efficient regularizations \cite{durrant2013random, alaoui2015fast, 
cannings2017random}.

Stochastic regularizations also exploit randomness for efficient
approaches to prevent overfit.
The prototypical example in neural networks is Dropout
\cite{srivastava2014dropout}. Dropout
modifies the network structure at each update within an SGD loop: it
removes units randomly
from the network during training and uses an approximate averaging procedure
across these ``thinned'' networks  during testing. Integrating random
perturbations within SGD gives a computationally cheap form of
ensembling \cite{bachman2014learning}. Dropout at the input layer can be viewed as data augmentation with random projections \cite{bouthillier2015dropout, vinh2016training}.

Another approach to tackle overfitting is by crafting models with suitable
inductive bias, for instance by imposing shared weights to capture
invariances of the data. Indeed, in a linear model or a fully-connected
layer of a neural network, the number of model weights increases with
the number of inputs and the
number of outputs. Convolutional neural networks (CNNs) circumvent this
difficulty with weight sharing. Rather than fitting one parameter per input pixel, CNNs re-use the same parameters by sliding a filter across
the input image. They typically use pooling layers that introduce
translation invariance and improve generalization
\cite{hinton2012improving}. Indeed, CNNs are very successful on natural images because
a cat should be modeled as the same object whether it is shifted to
the left or to the right. Such invariances also hold for text processing
\cite{kalchbrenner2014convolutional}, but not for brain activation
images. They display meaningful structure that is specific to given
features, \emph{i.e.} brain locations.
Non-translation-invariant problems require a departure from CNNs even in
computer vision, \emph{eg} with pixel-specific filters \cite{ren2015shepard}. 

\subsection{Proposed approach}
\begin{figure}[!t]
\includegraphics[width=\linewidth,trim=0pt .5cm 0pt 1.5cm,clip]{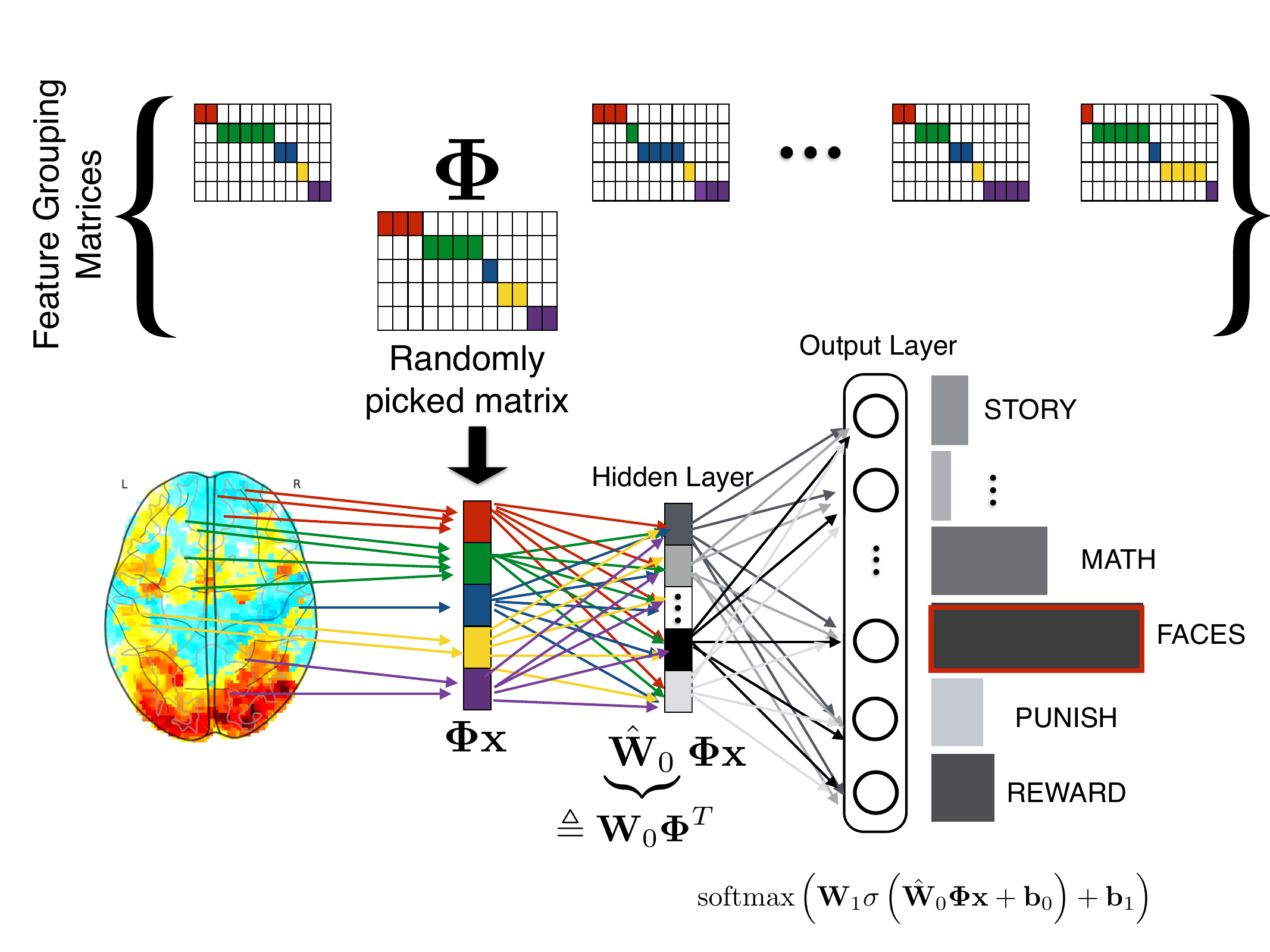}
\caption{\textbf{Illustration of the proposed approach}: Forward propagation of a neural network with a single hidden layer using \textit{feature grouping} during training. The parameters of the neural network to be estimated are $\bW_0, \bb_0, \bW_1, \bb_1$. A bank of feature grouping matrices are pre-generated where each matrix is calculated from a sub-sample of the training test. At each SGD iteration, a feature grouping matrix is sampled from the bank of pre-generated matrices. The gradient is then computed with respect to $\hat{\mathbf{W}}_0$ to update $\bW_0$ in backpropagation.}
\label{fig:figure_full_network}
\end{figure}

In this study, we use feature grouping to develop a
computationally efficient stochastic regularization approach for data
with a general dependency structure across the features. Our
algorithm relies on a bank of \textit{feature grouping} matrices to group the
features for training. These feature grouping matrices are adapted to the
data, but they can be pre-computed outside of the optimization loop
for computational efficiency.
Optimization is performed by a stochastic gradient descent (SGD),
sampling a matrix from the bank during forward propagation 
to project the features into a reduced representation.
The gradient is computed in this low-dimensional space. In order
to update the weights during back propagation, we project the gradient back to the
original feature dimension.  
This procedure results in weights expressed on
the original feature dimension (brain voxels for neuroimaging; pixels for image processing), that can be used
at test time.
%
As the projection matrices are sparse by design, projection is fast and adds only a small computational cost. On the other hand, gradients can be computed more efficiently in the lower-dimensional space. 
When applied to neural networks, our algorithm is suitable for the input layer only, because it relies on pre-computed projection matrices. These matrices depend on the values of the features, which do not change during training, unlike inputs to intermediate layers.
Standard regularization techniques should be applied to the subsequent layers. 
We target high-dimensional problems: the input layer typically has more
parameters than intermediate layers, and therefore calls for dedicated
regularization.
Figure \ref{fig:figure_full_network} illustrates our approach for a neural network with a single hidden layer.

The feature grouping approach we employ is a linear-time agglomerative
clustering scheme, Recursive Nearest Agglomeration (ReNA) proposed by
\citet{hoyos2019recursive}. 
ReNA is similar to the simple linear iterative
clustering (SLIC) algorithm \cite{achanta2012slic} used to produce super-pixels
in computer vision applications. 
The advantages of using a fast averaging procedure are two-fold: (i)
it has a denoising effect on structured signals; and (ii) it reduces the
dimension of signals with computation time linear in the feature dimension.

In the following sections, we provide details of the computational complexity of our approach. We also provide theoretical implications for generalized linear models.
We demonstrate the success of our approach in noisy and small-sample
settings by applying it to fully-connected multi-layer perceptrons (MLPs)
and logistic regressions on
the Olivetti faces dataset \cite{hopper1992orl}, and a publicly
available task fMRI data set from the Human Connectome Project
\cite{van2013wu}. In both cases, our approach outperforms
$\ell_2$ regularization and dropout applied to the same models, as well
as CNNs with dropout. Note that it cannot be combined with CNNs as the structured projection removes the redundant topography which convolutions exploit.
%
%
Experimental results demonstrate that feature grouping outperforms other methods by the greatest margin when the data size is limited and when the data are contaminated with noise.

\section{Model}

We consider supervised-learning settings. Let $\mathbf{x}
\in \mathbb{R}^p$ a feature vector with $y \in \mathbb{R}$
the corresponding target.
The model is a function
$f: \mathbb{R}^p \rightarrow \mathbb{R}$ with parameters
$\boldsymbol{\Theta}$.
These parameters are estimated by minimizing the
empirical risk over training samples $\left(\bx_i, y_i \right)$ for $i
\in \left\{1, \cdots, n \right\}$ such that: \be
\hat{\boldsymbol{\Theta}} = \arg \min_{\boldsymbol{\Theta}}
\frac{1}{n} \sum_{i=1}^n L\left(f (\bx_i; \boldsymbol{\Theta}), y_i
\right)
\label{eqn:main_erm}
\ee 
where $L$ is the cost per sample. For neural networks with an MLP architecture, the
parameter set is $\boldsymbol{\Theta} = \left\{ \bW_0,
\bb_0, \bW_1, \bb_1, \cdots, \bW_H,\bb_H \right\}$ where $H$ denotes
the number of hidden layers, $\bW_i$ represents the weights
and $\bb_i$ represents the bias at the $i$-th layer.

\subsection{Dimensionality reduction by feature grouping}
We assume that $\bx$ represents high-dimensional data with a strong
spatial structure as with fMRI data (where $p \sim 10^5-10^6$).
Reducing the dimensionality of these signals reduces memory
requirements and speeds up training steps.
This reduced representation helps training if the signal present in 
$\bx$ is preserved. This can be achieved by capturing the signal
structure in the dimensionality reduction.
Structure-aware dimensionality reduction is indeed known to be useful
for neuroimaging data \cite{mwangi2014review}.

We use a data-driven feature averaging approach, ReNA. The features are clustered, and their values are replaced with a single value for each cluster. %
Let
$\boldsymbol{\Phi} \in \mathbb{R}^{k \times p}$ be the dimensionality reduction 
matrix that
projects the data to a lower-dimensional space, with $k \ll p$.
The clusters are a partition of the features $\mathcal{P} = \left\{ \mathcal{C}_1,
\mathcal{C}_2, \cdots, \mathcal{C} _k\right\}$,
 where $\mathcal{C}_q$ is the set of indices that belong to
cluster $q$ and $\mathcal{C}_q \cap \mathcal{C}_l = \emptyset$ for $q
\neq l$. 
Approximation on the $q$-th cluster can be written as:
$\left( \boldsymbol{\Phi} \bx \right)_q = \alpha_q \sum_{j \in
  \mathcal{C}_q} \bx_j$, where $\alpha_q = 1 / \sqrt{\text{card}( \mathcal{C}_q
  )}$ is a constant for cluster $q$
  chosen to make $\boldsymbol{\Phi}$ an orthogonal 
matrix. 
$\boldsymbol{\Phi} \bx \in \mathbb{R}^k$ is the projected, or reduced, version of
$\bx$ and $\boldsymbol{\Phi}^T \boldsymbol{\Phi} \bx$ is a piecewise constant
approximation of $\bx$.

We use the ReNA clustering algorithm to obtain the projection matrices
$\boldsymbol{\Phi}$. ReNA is a graph-constrained clustering: when the
graph represents the dependencies between the features of the signal,
feature grouping with 
ReNA has been shown to have a denoising effect which improves subsequent
analysis \cite{hoyos2019recursive}.
The algorithm starts with $p$ clusters, one per feature.
Clusters are then recursively merged until the desired number of clusters remain. 
Merging is achieved by a greedy graph cutting algorithm. For data on a
grid, as with image data, the initial graph connects 
pixels or voxels to their neighbors, with edge weights determined by the
data. 
A new graph is constructed to express the connectivity after merging
features, and the process is repeated.
Though we use ReNA, our framework can employ any
clustering algorithm. The benefits of ReNA are that it is a fast
structured clustering algorithm that leads to good signal approximations.

\subsection{Stochastic Regularizer with Feature Grouping}
We now describe our algorithm. 
First, we generate a bank of feature grouping matrices $\varPhi = \left\{ \boldsymbol{\Phi}^{(1)},  \boldsymbol{\Phi}^{(2)}, \cdots,  \boldsymbol{\Phi}^{(b)}  \right\}$ using ReNA. Each $ \boldsymbol{\Phi}^{(i)}$ is generated using $r$ samples from the training data set selected randomly with replacement. 
Then we begin the SGD loop for model training. At each iteration, which consists of a gradient calculation and a weight update, we sample a random $\boldsymbol{\Phi}^{(i)}$
from the bank $\varPhi$. We use $\boldsymbol{\Phi}^{(i)}$ to project the training samples onto a lower dimensional space, and compute gradients in this lower dimensional space.
This operation affects only the weight matrix in the input layer of the neural network, while subsequent weights and all biases are treated in a standard way. 
Instead of computing the gradient with respect to the $h \times p$
dimensional matrix $\bW_0$, where $h$ is the number of units in the first
hidden layer, we compute the gradient with respect to the $h \times k$
dimensional weight matrix, called $\hat{\bW}_0 \overset{def}{=}\bW_0 \boldsymbol{\Phi}^{T}$. Computational operations with $\hat{\bW}_0$ are much cheaper  than those with ${\bW}_0$ because $k \ll p$.
 
During the update of $\bW_0$, we project the gradient back to the
original space. This operation can be interpreted as using $\bW_0
\boldsymbol{\Phi}^T \boldsymbol{\Phi}$ as a weight matrix instead of
$\bW_0$. Since $\boldsymbol{\Phi}^T \boldsymbol{\Phi} \bx$ is an
approximation of $\bx$, it is equivalent to deriving the weight matrix
$\bW_0$ from the approximation of the input. Feature grouping acts as a stochastic regularizer by forcing the model to learn from these approximated inputs.

\setlength{\textfloatsep}{8pt}
\begin{algorithm}[!t]
\caption{Training of a Neural Network with Feature Grouping as a Stochastic Regularizer}
\begin{algorithmic}[1]
  \small
  \REQUIRE Learning Rate $\eta$
  \REQUIRE Initial Parameters for $H$ layers \\\qquad $\boldsymbol{\Theta} \triangleq \{ \bW_0, \bb_0, \bW_1, \bb_1, \cdots, \bW_H, \bb_H \}$
  \ENSURE Generate a bank of feature grouping matrices where each is generated by randomly sampling $r$ samples from the training data set with replacement \\\qquad$\varPhi = \left\{ \boldsymbol{\Phi}^{(1)},  \boldsymbol{\Phi}^{(2)}, \cdots,  \boldsymbol{\Phi}^{(b)}  \right\}$ 
  \WHILE{stopping criteria not met}
       \STATE Sample a minibatch of m samples from the training set $\{\bx^{(1)}, \cdots, \bx^{(m)} \}$ with corresponding labels $y^{(i)}$
       \STATE Sample $\boldsymbol{\Phi}$ from the bank $\varPhi$.
       \STATE Define $\boldsymbol{\Xi} \triangleq \left\{ \hat{\mathbf{W}}_0, \mathbf{b}_0, \bW_1, \bb_1, \cdots, \bW_H, \bb_H  \right\}$ where $\hat{\bW}_0 \triangleq \bW_0 \boldsymbol{\Phi}^T$.
       \STATE Compute gradient estimate: \\\qquad$\bg \gets \frac{1}{m} \nabla_{\boldsymbol{\Xi}} \sum_i \mathcal{L}\left( f(\boldsymbol{\Phi}\bx^{(i)}; \boldsymbol{\Xi}), y^{(i)} \right)$
       \STATE Apply updates: 
       \begin{itemize}
       \item $\mathbf{W}_0 \gets \mathbf{W}_0 - \eta  \bg_{\mathbf{w}_0}
\boldsymbol{\Phi}$ \\\quad where $\mathbf{g}_{\mathbf{w}_0} \triangleq \frac{1}{m} \nabla_{\hat{\mathbf{W}}_0} \sum_i \mathcal{L}\left( f(\boldsymbol{\Phi}\bx^{(i)}; \boldsymbol{\Xi}), y^{(i)} \right)$
        \item $\mathbf{b}_j \gets \mathbf{b}_j - \eta  \bg_{b_j}$
\\\quad where $\mathbf{g}_{b_j} \triangleq \frac{1}{m} \nabla_{\mathbf{b}_j} \sum_i \mathcal{L}\left( f(\boldsymbol{\Phi}\bx^{(i)}; \boldsymbol{\Xi}), y^{(i)} \right)$ \\\quad for $j \in \left\{0, \cdots, H \right\}$
\item $\mathbf{W}_j \gets \mathbf{W}_j - \eta  \bg_{\mathbf{w}_j}$ \\\quad where $\mathbf{g}_{\mathbf{w}_j} \triangleq \frac{1}{m} \nabla_{{\mathbf{W}}_j} \sum_i \mathcal{L}\left( f(\boldsymbol{\Phi}\bx^{(i)}; \boldsymbol{\Xi}), y^{(i)} \right)$ \\\quad for $j \in \left\{1, \cdots, H \right\}$
 \end{itemize}
  \ENDWHILE
\end{algorithmic} \label{alg:rena_with_sgd}
\end{algorithm}

We describe the resulting estimator for training a neural network with $H$ layers in Algorithm \ref{alg:rena_with_sgd}. Since the weights ${\bW}_0$ we learn match the original feature dimension, the grouping matrices can be discarded after training completes, and no special procedure is needed at test time.

\subsection{Interpretation of the proposed approach}
\setlength\abovedisplayskip{7pt plus2pt minus4pt}
\setlength\belowdisplayskip{7pt plus2pt minus4pt}
With randomized feature grouping matrices in the SGD, we are effectively computing the parameters such that:
\be
\hat{\boldsymbol{\Theta}} = \arg \min_{\boldsymbol{\Theta}} \frac{1}{n} \sum_{i=1}^n \mathbb{E}_{\boldsymbol{\Phi}} \left[L\left( f(\boldsymbol{\Phi}^T \boldsymbol{\Phi} \bx_i; \boldsymbol{\Theta}), y_i \right) \right] \label{eqn:main_optimization}
\ee
instead of Equation \ref{eqn:main_erm}. We investigate the effect of this
approach for generalized linear models (GLM) as used by \citet{wager2013dropout}
to uncover dropout's properties. Here
$\boldsymbol{\Theta} = \left\{ \boldsymbol{\beta} \right\}$, where
$\boldsymbol{\beta}$ is a vector of parameters.
The generalized linear model framework
models the response $y$ given a feature
vector $\bx$ and the model parameter $\boldsymbol{\beta}$ as:
\be
p \left( y \mid \bx ;  \boldsymbol{\beta} \right) \triangleq h(y) \exp
\biggl( y  \bx^T \boldsymbol{\beta} - A\left( \bx^T \boldsymbol{\beta}
\right) \biggr) 
\ee
where $h(y)$ is a quantity independent of $\bx$ and $\boldsymbol{\beta}$;
and $A(.)$ is the log-partition function which is equivalent to $\| \bx^T
\boldsymbol{\beta} \|^2$ for a least squares regression or Gaussian model.

We now separate $\boldsymbol{\Phi}$ in two terms:
$ {\boldsymbol{\Phi}}^T {\boldsymbol{\Phi}} =
{\boldsymbol{\Omega}} +  {\boldsymbol{\Delta}}$ where
$\boldsymbol{\Omega} = \mathbb{E}[\boldsymbol{\Phi}^T
{\boldsymbol{\Phi}}]$ is the deterministic term and ${\boldsymbol{\Delta}}$ is zero-mean noise term such that $\mathbb{E} \left[{\boldsymbol{\Delta}}\right] = \mathbf{0}$.
$\boldsymbol{\Omega}$ captures the commonalities across multiple
realizations of ReNA. As these are shaped by the feature graph used to
impose structure, $\boldsymbol{\Omega}$ typically resembles a graph smoothing
operator. 
The sum in Equation \ref{eqn:main_optimization} can then be written as:
\bea
\lefteqn{ \sum_{i=1}^n  \mathbb{E}_{\boldsymbol{\Phi}} \left[ L\left(f \left( \boldsymbol{\Phi}^T \boldsymbol{\Phi} \bx_i; \boldsymbol{\Theta} \right), y_i \right) \right]} \\
&=&  \sum_{i=1}^n -y_i \mathbf{x}_i^T \boldsymbol{\Omega} \boldsymbol{\beta} + \mathbb{E}_{\boldsymbol{\Phi}} \left[ A \left( \bx_i^T \left( \boldsymbol{\Omega} +  \boldsymbol{\Delta} \right) \boldsymbol{\beta} \right) \right] \label{eqn:before_quadratic} 
\eea
 We apply second-order Taylor approximation to the term $ A\left( \bx^T \left( \boldsymbol{\Omega} +  \boldsymbol{\Delta} \right) \boldsymbol{\beta} \right)$ around $\bx^T \boldsymbol{\Omega} \boldsymbol{\beta}$ as a standard quadratic approximation also used by \citet{bishop1995training, rifai2011adding,wager2013dropout} and take the expectation:
 \bea
 \lefteqn{\mathbb{E}_{\boldsymbol{\Phi}} \left[ A\left(\bx^T ( \boldsymbol{\Omega} + \boldsymbol{\Delta})  \boldsymbol{\beta} \right) \right] \approx }\nonumber \\
 & & A\left(\bx^T \boldsymbol{\Omega} \boldsymbol{\beta} \right) + \frac{1}{2} A''\left(\bx^T \boldsymbol{\Omega} \boldsymbol{\beta} \right) \mathbb{E}_{\boldsymbol{\Phi}}\left[ \| \bx^T \boldsymbol{\Delta} \boldsymbol{\beta} \|^2  \right] 
 \eea
The first-order term $\mathbb{E}_{\boldsymbol{\Phi}} \left[ A'\left(\bx^T \boldsymbol{\Omega} \boldsymbol{\beta} \right) \bx^T \boldsymbol{\Delta} \boldsymbol{\beta} \right]$ vanishes because $\mathbb{E} \left[{\boldsymbol{\Delta}}\right] = \mathbf{0}$.
 Substituting this into Equation \ref{eqn:before_quadratic} gives:
 \bea
\lefteqn{ \sum_{i=1}^n  \mathbb{E}_{\boldsymbol{\Phi}} \left[ L\left( f \left( \boldsymbol{\Phi}^T \boldsymbol{\Phi} \bx_i; \boldsymbol{\Theta} \right), y_i \right) \right] \approx } \nonumber \\
& & \sum_{i=1}^n  \underbrace{-y_i \mathbf{x}_i^T \boldsymbol{\Omega}  \boldsymbol{\beta} + A\left(\bx_i^T \boldsymbol{\Omega} \boldsymbol{\beta} \right)}_{L(\boldsymbol{\Omega} \bx_i, y_i; \boldsymbol{\beta})}  \nonumber \\
&  & + \frac{1}{2} \underbrace{\sum_{i=1}^n A''\left(\bx_i^T \boldsymbol{\Omega} \boldsymbol{\beta} \right)  \mbox{Var}_{\boldsymbol{\Phi}}\left[ \bx_i^T \boldsymbol{\Phi}^T \boldsymbol{\Phi}  \boldsymbol{\beta} \right]}_{\triangleq R\left( \boldsymbol{\beta} \right)} 
 \eea
 The above equation shows that our cost function consists of two terms:
(i) the loss on the smoothed input $\boldsymbol{\Omega}\,\bX$ (ii) a regularization cost $R\left( \boldsymbol{\beta} \right)$. 
It is known that the term $A''\!\left(\bx_i^T  \boldsymbol{\beta} \right)$ corresponds to the variance of $y_i$ given $\bx_i$ under GLM settings
\cite{mcculloch2001generalized}. Hence $A''\!\left(\bx_i^T
\boldsymbol{\Omega} \boldsymbol{\beta} \right)$ is the variance of the
model given the smoothed input features $\boldsymbol{\Omega}\bx_i$. Note that this term is constant for linear regression and equivalent to $p_i (1 - p_i)$ where $p_i = 1 / (1 - \exp{(- \bx_i^T \boldsymbol{\Omega} \boldsymbol{\beta})})$ with feature grouping for logistic regression.

The term $ \mbox{Var}_{\boldsymbol{\Phi}}\left[ \bx_i^T \boldsymbol{\Phi}^T \boldsymbol{\Phi}  \boldsymbol{\beta} \right]$ corresponds to the variance of the estimated target due to the randomization introduced by stochastic regularizer. Using the definition $\boldsymbol{\Phi}^T\boldsymbol{\Phi} = \boldsymbol{\Omega} + \boldsymbol{\Delta}$ and symmetricity of $\boldsymbol{\Delta}$, it reduces to:
\bea
\mbox{Var}_{\boldsymbol{\Phi}}\left[ \bx_i^T \boldsymbol{\Phi}^T \boldsymbol{\Phi}  \boldsymbol{\beta} \right] &=& \mbox{Var}_{\boldsymbol{\Phi}}\left[ \bx_i^T \boldsymbol{\Delta}  \boldsymbol{\beta} \right] \nonumber \\
&=& \boldsymbol{\beta}^T \mathbb{E}\left[ \boldsymbol{\Delta} \bx_i \bx_i^T  \boldsymbol{\Delta} \right] \boldsymbol{\beta}
\eea

If we used a $\boldsymbol{\Phi}$ matrix corresponding to a dropout on the
input layer,
randomly masking features, we would have $\boldsymbol{\Omega} = \mathbf{I}$
and $\boldsymbol{\Delta}$ a diagonal matrix where each $i$-th diagonal
term has $\mathbb{E} \left[ \Delta_i \right] = 0$ and $\mathbb{E} \left[
\Delta_i^2 \right] = \delta / (1 - \delta)$ where $\delta$ is the dropout
probability. Assuming $\mathbb{E}\left[\Delta_i \Delta_j \right] = 0$ for $i \neq j$, it can be written as:
\be
 \mbox{Var}_{\boldsymbol{\Phi}}\left[ \bx_i^T \boldsymbol{\Phi}^T \boldsymbol{\Phi}  \boldsymbol{\beta} \right] = \frac{\delta}{1-\delta} \sum_{j=1}^p x_{ij}^2 \beta_j^2
\ee
where $x_{ij}$ is the $j$-th entry of $\bx_i$. For linear regression, this is equivalent to ridge regression after orthogonalizing the features. 

However, for feature grouping, the matrix $\mathbb{E}\left[
\boldsymbol{\Delta} \bx_i \bx_i^T  \boldsymbol{\Delta} \right]$ rescales
the feature vector $\bx_i$ by the variance of the cluster membership for
each feature. 
For instance, if a feature consistently appears in a certain cluster, then 
the membership variance for this feature will be low. If, on the other hand, a feature appears in a certain cluster only in half of the samples, then the variance will be high and will have a large weight in penalty term $\mbox{Var}_{\boldsymbol{\Phi}}\left[ \bx_i^T \boldsymbol{\Phi}^T \boldsymbol{\Phi}  \boldsymbol{\beta} \right]$. 
As the clusters in $\boldsymbol{\Phi}$ are obtained from bootstrap replicates 
of the data,
this penalty term captures the local spatial stability of the data.

\begin{figure}[!t]
  \centering
  \mbox{%
  \subfigure[$\boldsymbol{\Phi}_1$]{\includegraphics[scale=0.17,trim={0
3.3cm 0 3.3cm},clip]{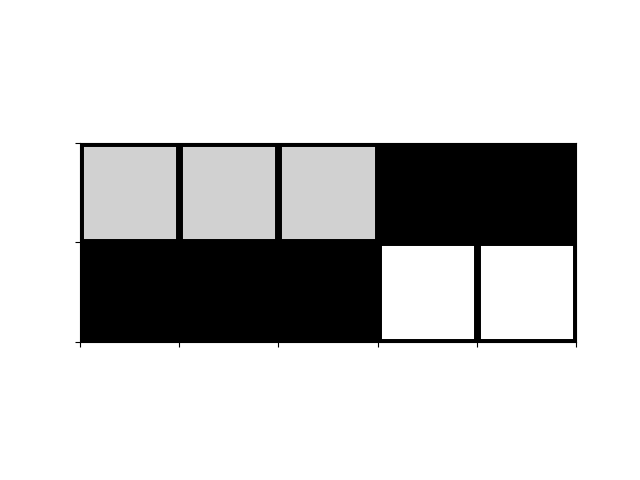} \label{fig:toy_phi1}} \hfill
    \subfigure[$\boldsymbol{\Phi}_2$]{\includegraphics[scale=0.17,trim={0
3.3cm 0 3.3cm},clip]{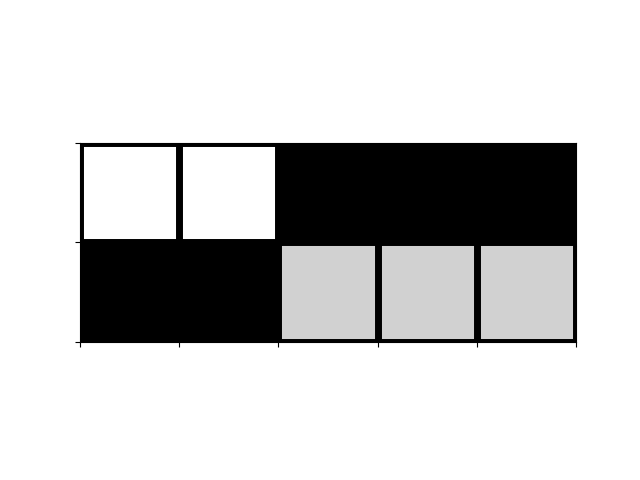} \label{fig:toy_phi2}}}
    \mbox{%
    \subfigure[$\boldsymbol{\Omega} \triangleq \mathbb{E}{[\boldsymbol{\Phi}^T
\boldsymbol{\Phi}]}$]{\includegraphics[scale=0.17,trim={0
1.2cm 0 1.2cm},clip]{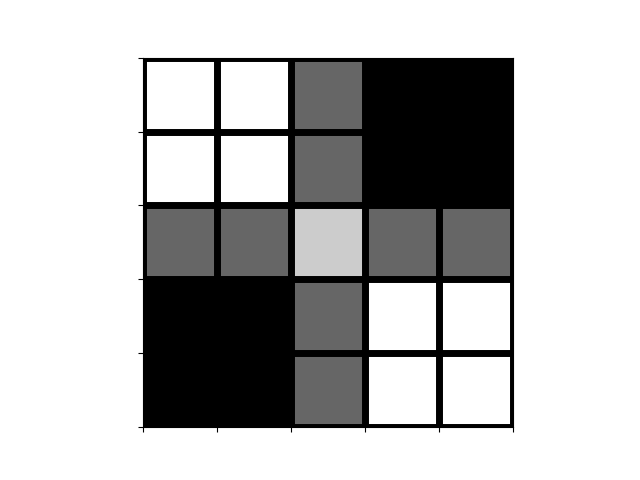} \label{fig:toy_omega}}\hfill
    \subfigure[ $\mathbb{E}{[\boldsymbol{\Delta}^T
\boldsymbol{\Delta}]}$]{\includegraphics[scale=0.17,trim={0 1.2cm 0 1.2cm},clip]{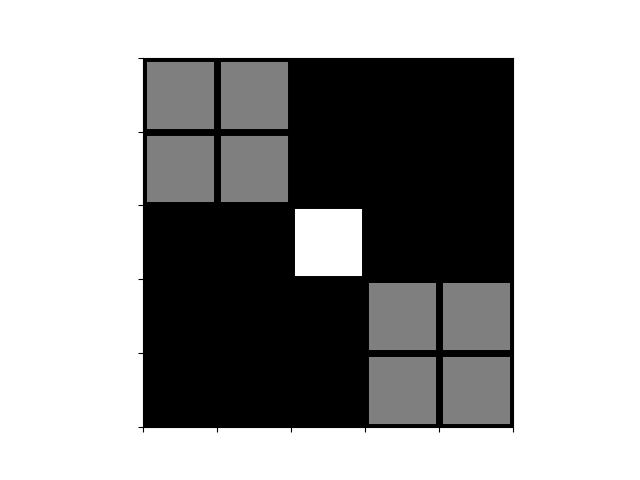} \label{fig:toy_delta}}
}
\caption{Visualization for a toy example where number of feature is $p=5$
and number of clusters if $k=2$. (a) and (b) The two feature-grouping
matrices $\boldsymbol{\Phi}_1$ and $\boldsymbol{\Phi}_2$ (c) Average of
$\boldsymbol{\Phi}_1$ and $\boldsymbol{\Phi}_2$; i.e.
$\boldsymbol{\Omega}$ (d) Variance of $\boldsymbol{\Phi}$, i.e. estimated
variance of $\boldsymbol{\Delta}$. $\boldsymbol{\Omega}$ indeed does
appear as a smoothing matrix, and $\mathbb{E}{[\boldsymbol{\Delta}^T
\boldsymbol{\Delta}]}$ captures the spatial homogeneity: it is large for
the central feature while the sides are more smoothed by
$\boldsymbol{\Omega}$ and stabilized by the edges.}
\label{fig:toy_example}
\end{figure}

Figure \ref{fig:toy_example} illustrates these terms with a toy example.
Our bank of feature-grouping matrices is made of 2 matrices, 
$\boldsymbol{\Phi}_1$ and $\boldsymbol{\Phi}_2$. Note that the third
feature appears with the first two features in matrix
$\boldsymbol{\Phi}_1$ whereas it appears with the last two features in
$\boldsymbol{\Phi}_2$. Figure \ref{fig:toy_omega} shows the average of
$\boldsymbol{\Phi}_i^T \boldsymbol{\Phi}_i$
which captures the general topography of the groups. Figure
\ref{fig:toy_delta} shows the variance of these two matrices which
captures the high variance of feature $3$. This way, our algorithm
penalizes the features that are more noisy via the term
$\mbox{Var}_{\boldsymbol{\Phi}}\left[ \bx_i^T \boldsymbol{\Phi}^T
\boldsymbol{\Phi}  \boldsymbol{\beta} \right]$ in $R(\boldsymbol{\beta})$
while used smoothed features $\boldsymbol{\Omega}$ in the optimized loss function $L \left(\boldsymbol{\Omega}\bx_i, y_i; \boldsymbol{\beta} \right)$ and the
regularized term $A''\left(\bx^T \boldsymbol{\Omega} \boldsymbol{\beta} \right)$ in $R(\boldsymbol{\beta})$.

\subsection{Computational Complexity}
The computational complexity of optimizing a given neural
network with the feature-grouping stochastic regularizer differs from the standard approach only for the parameter $\bW_0$. Learning the rest of
the parameters $\left\{\bb_0, \bW_1, \bb_1, \cdots, \bb_H, \bW_H
\right\}$ is unchanged.
Therefore, it is sufficient to compare performance for logistic regression where the size of $\bW_0$ is $l \times p$ instead of $h \times p$ where $l$ is the total number of classes.
The  computational complexity of logistic regression, solved with the stochastic regularizer using feature grouping breaks down in four parts: \textit{(i)} computation of the bank of $\boldsymbol{\Phi}$ matrices (Step 1 in
Algorithm \ref{alg:rena_with_sgd}; ) \textit{ii)} multiplication by $\boldsymbol{\Phi}$ in summation in Step 6;  \textit{(iii)} computation of gradient in Step 6; and \textit{iv)} update in Step 7.
%

The computational complexity of computing each $\boldsymbol{\Phi}$ using
ReNA is $\mathcal{O} \left( r p \log\left(p/k \right)\right)$
\cite{hoyos2019recursive} where $r$ is the number of samples used, $p$ is the number of features and $k$ is the number of clusters. Since the bank has $b$ such matrices, the total computational complexity of computing the bank is $\mathcal{O} \left( b r p \log\left(p/k \right)\right)$. This is a constant factor independent of the number of iterations. 

Computing $\boldsymbol{\Phi}\bx$, where $\boldsymbol{\Phi}$ is of dimension $p \times k$ would be $\mathcal{O}\left( kp \right)$. However, as $\boldsymbol{\Phi}$ is sparse, this reduces to $\mathcal{O}\left( p \right)$. Since there are $m$ samples in a minibatch, the total computational complexity is $\mathcal{O}\left( mp \right)$. %
Computational complexity of gradient computation for a row of $\hat{\bW}_0$ for a given sample $\bx^{(i)}$ is $\mathcal{O} \left( k \right)$. 
Computing the gradient across all rows and samples in a minibatch has complexity $\mathcal{O} \left( lmk \right)$, $l$ being the number of tasks.
Updating one row of $\bW_0$ requires right multiplying the gradient with respect to $\hat{\bW}_0$ by $\boldsymbol{\Phi}^T$ which would be $\mathcal{O}\left(kp \right)$, but due to the sparse structure of $\boldsymbol{\Phi}$, it reduces to $\mathcal{O}\left( p \right)$.
As there are $l$ rows, the total computational complexity for update is $\mathcal{O}\left( lp \right)$. 

Projection, gradients computation, and update are done for each epoch, so the computational complexity for the full iteration would be $\mathcal{O} \left( T\,m\,p + T\,l\,m\,k + T\,l\,p \right)$ where $T$ is the total number of iterations. Hence the total computational complexity of logistic
regression with feature grouping using ReNA can be written as:
\be
\mathcal{O} \left( b\, r\, p \log\left(p/k \right) + T\,m\,p + T\,l\,m\,k + T\,l\,p \right)
\ee
which is linear in the dimension of the input size $p$ and number of classes $l$. Computational complexity of standard logistic regression, on the other hand, is $\mathcal{O}\left(T \,l\,m\,p \right)$.
 
\section{Experiments}
We presented a regularization algorithm that relies on feature grouping.
Our approach can be easily integrated into fully-connected feedforward
neural networks. In order to validate the effectiveness of our algorithm
and compare it with conventional approaches we experiment in noisy and low sample size settings on face (Olivetti) and neuroimaging (HCP) datasets:

\textbf{Olivetti Faces:} The Olivetti dataset consists of grayscale $64 \times 64$ face images
from $40$ subjects \cite{hopper1992orl}. 
For each subject, there are $10$
different images with varying lighting and facial expressions. The target
class for this data set is the identity of the individual whose picture
was taken. We randomly split the data into test and train such that the
test dataset has $132$ samples and the training dataset, $268$ samples.
As the faces are well centered, the data has a strong non-translation-invariant structure.

\textbf{HCP:} The Human Connectome Project (HCP) has released a large
openly-accessible fMRI dataset. Here we use task fMRI that includes seven tasks: 1. Working Memory, 2. Gambling, 3. Motor, 4. Language 5. Social Cognition, 6. Relational Processing, and 7. Emotion Processing. 
These tasks have been chosen to map different brain systems. The dataset includes $500$ different subjects with images registered to the standard MNI atlas. For a given subject and task, a GLM was fitted to each fMRI dataset \cite{barch2013function}. Then volumetric contrasts of parameter estimate (COPE) were computed to assess differences between different task components, resulting into brain maps. We use $20$ different contrasts as described in Table
\ref{table:tasks}.  fMRI data are sampled in a common space of $91
\times 109 \times 91$ with $2$mm isotropic voxels. We transformed 3D
data into 1D arrays of size $p=270\,806$ for our supervised
classification algorithms.
Our goal is to classify $20$ cognitive contrasts given $p=270\,806$
features. The test dataset includes $1\,964$ samples with at least $95$
samples from each target class whereas the training set has $7\,785$
samples.

\textbf{HCP - small:}
In order to perform fast experimentation, we use a smaller number of classes and voxels from the HCP data set.
We select $8$ different contrasts from tasks: 1. Working Memory, 2. Gambling, 3. Relational, 4. Emotion, and 5. Social as described in Table
\ref{table:small_tasks} that are harder to classify. fMRI data are resampled to a common space of $46 \times 55 \times 46$ with $4$mm isotropic voxels. We transformed 3D data into 1D arrays of size $33,854$. Our goal is to classify $8$
cognitive contrasts given $33,854$ features.
The test data includes $791$ samples with at least $97$ samples from
each target class whereas the training set has $3052$ samples. 

\subsection{Architectures}
We used three typical machine learning architectures in our experiments: (i) logistic regression (ii) multilayer perceptron (MLP) with a single hidden layer of size $256$, and (iii)  convolutional neural network (CNN) \cite{lecun2015lenet} that consists of two convolutional layers, two sub-sampling (pooling) layers and two fully connected layers. Convolutional layers for the Olivetti data set used $5 \times 5$ convolutions with stride $1$ and the sub-sampling layers are $2 \times 2$ max pooling layers. Convolutional layers for the HCP and HCP-small data sets used $7 \times 7$ and $5 \times 5$ convolutions with stride $2$ in the first and second layers, respectively and the sub-sampling layers are $2 \times 2$ max pooling layers. ReLU activation functions are used both in MLP and CNN.

\subsection{Training}
We use the standard SGD algorithm with learning rate $0.01$ for the
Olivetti dataset and $0.05$ for the HCP dataset. We use a cross entropy
loss. We ran experiments for logistic regression long enough (200 epochs for Olivetti and 500 epochs for HCP and HCP-small) to guarantee convergence. We applied \textit{early stopping} on MLP and CNN architectures when the validation loss stopped improving in $10$ (also known as patience parameter) subsequent epochs. We repeated each experiment with $10$ different random initializations.

\subsection{Parameter tuning for regularizers}
We vary the regularization parameter for $\ell_2$ from $10^{-7}$ to $10$
with a grid of factors of 10, and the dropout probability parameter for  dropout from $0.1$ to $0.7$ with a grid of $0.2$ for logistic regression and MLP architectures. We used only dropout with dropout probability $0.5$ for CNNs. 

\begin{figure}[!t]
\mbox{
\hspace{-.61cm}
\subfigure{\includegraphics[scale=0.129]{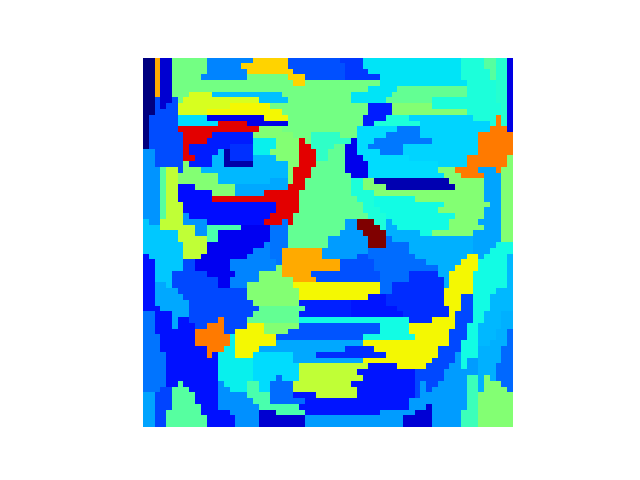} \label{fig:phis_0}} 
\hspace{-.61cm}
\subfigure{\includegraphics[scale=0.129]{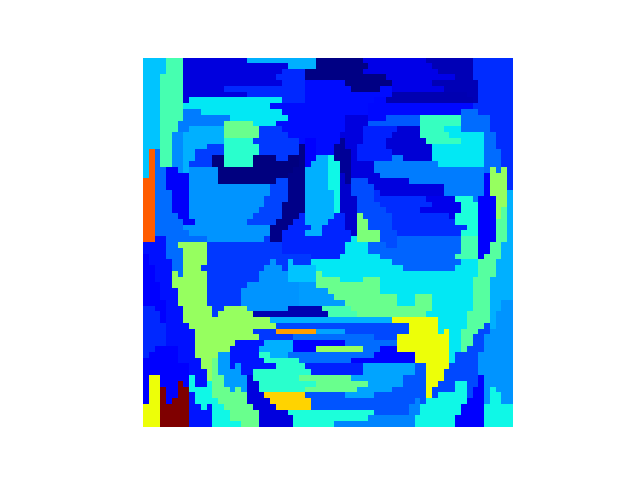} \label{fig:phis_1}}
\hspace{-.61cm}
\subfigure{\includegraphics[scale=0.129]{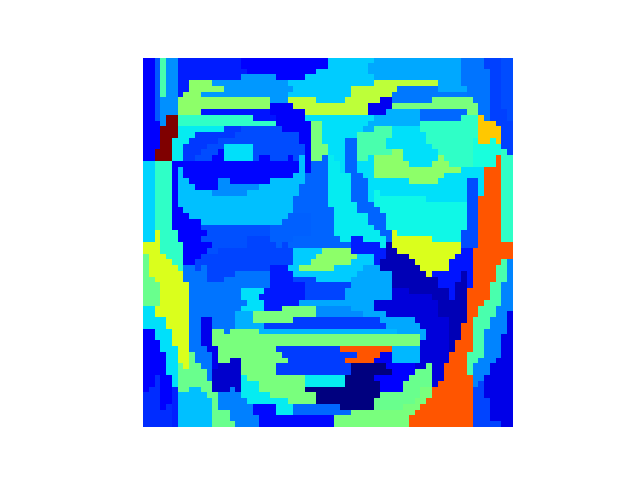} \label{fig:phis_2}}  
\hspace{-.61cm} 
\subfigure{\includegraphics[scale=0.129]{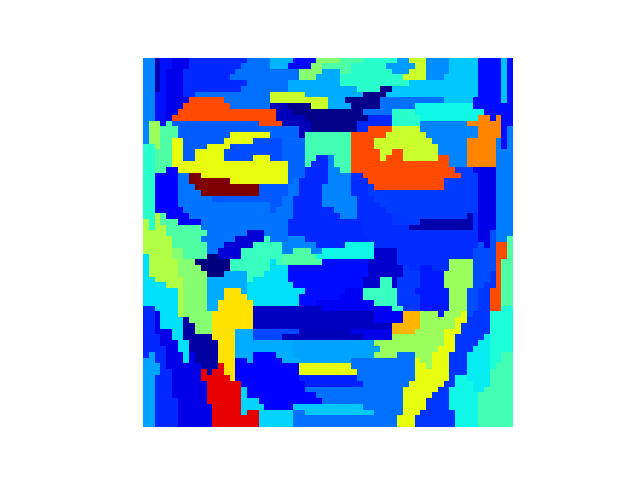} \label{fig:phis_3}}
\hspace{-.61cm}
\subfigure{\includegraphics[scale=0.129]{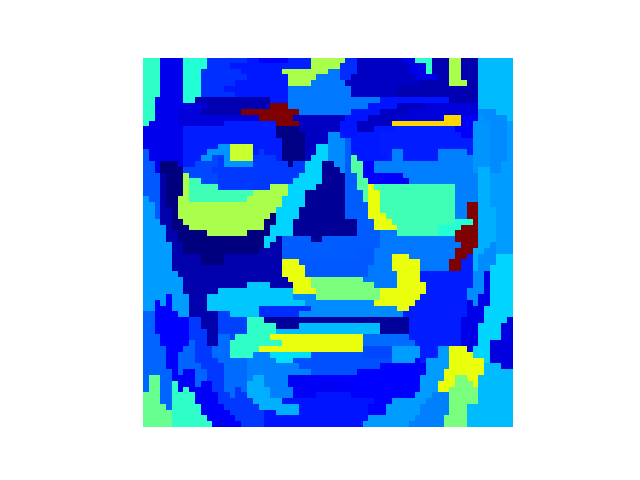} \label{fig:phis_4}}}
\caption{Sampled feature grouping matrices from the bank. Each matrix is computed using randomly selected $50$ samples from the Olivetti faces training data set. Note that each row of $\boldsymbol{\Phi}$ matrix is reshaped to the size of the image and then all rows are overlayed for visualization purposes.}
\label{fig:phis}
\end{figure}

For our approach, each feature grouping matrix $\mathbf{\Phi}$ is computed over $r = 50$ randomly picked
samples from the training data set for $k$ clusters, where $k$ is  set  to 10 \% of the total number of features. For each epoch during training, a feature grouping matrix was randomly picked from the bank of $b=100$ matrices. Figure \ref{fig:phis} shows samples of feature grouping matrices from the bank. Empirically, feature grouping
regularization is not very sensitive to the choice of $b$ and
$r$ (Table \ref{table:rb_accuracy}). For MLP, we also combine
feature grouping with dropout at intermediate layers to regularize them.

\subsection{Computational details}
We use Python 3.6 for implementation \cite{oliphant2007python} using open-source libraries PyTorch \cite{paszke2017pytorch}, scikit-learn \cite{pedregosa2011scikit}, NiBabel \cite{brett2016nibabel}, nilearn \cite{abraham2014machine}, joblib \cite{varoquaux2009joblib} and NumPy \cite{walt2011numpy}. Experiments using Olivetti and HCP-small data sets are run using Nvidia GeForce GTX 1060 and 16GB RAM. We use a computer with 128 GB RAM without a GPU to run experiments for HCP data set because the memory of the machine with the GPU could not hold a single data sample. Furthermore, we had to use a much smaller batch size ($10$) for CNNs as opposed to logistic regression and MLPs ($200$) because CNNs demand more memory. Our experiments for the Olivetti data set can be \textit{reproduced via the code provided in \url{https://github.com/sergulaydore/Feature-Grouping-Regularizer}.}

\begin{table}[!t]
\small
\begin{tabular}{|p{.15cm} |p{.6cm} |p{2.05cm}|p{1.62cm}|p{1.62cm}|}
\hline
\multirow{2}{*}{} &
\multirow{2}{*}{\tiny{\textbf{\!MODEL}}} & \multirow{2}{*}{\tiny{\textbf{REGULARIZER}}} & \multicolumn{2}{c|}{\textbf{\tiny{\textbf{TEST ACCURACY (\%)}}}} \\ 
\cline{4-5}
& & & \tiny{\textbf{OLIVETTI}}  & \tiny{\textbf{HCP-small}}\\
\hline
\multirow{9}{8em}{\rotatebox[origin=c]{90}{No noise}} & {\multirow{3}{*}{LR} } & None & $85.23 \pm 0.70$ & $86.80 \pm 0.18$ \\
& & Best $\ell_2$ & $86.52 \pm 1.01$ & $87.22 \pm  0.12$ \\ 
& & Best dropout & $85.76 \pm 0.88$ & $87.35 \pm 0.15$ \\ 
& & feature grouping & {$\mathbf{86.52 {\pm} 0.68}$} & {$\mathbf{87.37 {\pm} 0.29}$} \\
\cline{2-5}
& {\multirow{3}{*}{MLP} } & None & ${85.38 \pm 1.10} $ & ${88.02 \pm 0.18} $ \\
& & Best $\ell_2$ & $87.73 \pm 0.81$ & {$\mathbf{88.31 {\pm} 0.14}$} \\ 
& & Best dropout & {$\mathbf{89.55 {\pm} 0.88}$}  & $87.72 \pm 0.13$ \\ 
&  & feature grouping & $85.45 \pm 1.08$ & $87.36 \pm 0.57$ \\
\cline{2-5}
& {\multirow{1}{*}{CNN} } & dropout, $p=.5$ & $ 83.56 \pm 1.43$ & $ 74.96 \pm 0.61$ \\
\cline{2-5}

\hline
\multirow{9}{8em}{\rotatebox[origin=c]{90}{Medium noise level} } & {\multirow{3}{*}{LR} } & None & $50.83 \pm 1.79 $ & $79.77 \pm 0.35$ \\
& & Best $\ell_2$& $51.06 \pm 1.16$  & $79.97 \pm 0.36 $ \\ 
& &  Best dropout & $52.20 \pm 1.21$ & $79.90 \pm 0.30$ \\ 
& &  feature grouping & {$\mathbf{80.00 {\pm} 0.83}$} & {$\mathbf{84.16 {\pm} 0.24}$} \\
\cline{2-5}
& {\multirow{3}{*}{MLP} } & None & $54.55 \pm 1.63$ & $76.94 \pm 0.20$ \\
& &  Best $\ell_2$ & $56.59 \pm 1.53$ & $80.66 \pm 0.22$ \\ 
& &  Best dropout& $61.82 \pm 1.14$ & $80.01 \pm 0.42$ \\ 
& &  feature grouping & {$\mathbf{80.91 {\pm} 1.02}$} & {$\mathbf{83.75 {\pm} 0.35}$} \\
\cline{2-5}
&  {\multirow{1}{*}{CNN} }  & dropout, $p=.5$ & $77.65 \pm 1.11$ & $63.94 \pm 1.27$ \\

\hline
\multirow{9}{8em}{\rotatebox[origin=c]{90}{High noise level}} &  {\multirow{3}{*}{LR} } & None & $22.27 \pm 0.54$  & $71.42 \pm 0.61$ \\
& & Best $\ell_2$ & $24.62 \pm 1.50$ & $71.76 \pm 0.43$ \\ 
& & Best dropout& $24.09 \pm 1.54$ & $72.10 \pm 0.48$ \\ 
& & feature grouping & {$\mathbf{64.55 {\pm} 1.77}$} & {$\mathbf{77.93 {\pm} 0.38}$} \\
\cline{2-5}
& {\multirow{3}{*}{MLP} } & None & $25.00 \pm 1.89$ & $62.92 \pm 0.40$ \\
& & Best $\ell_2$& $28.56 \pm 2.09$ & $69.13 \pm 0.32$ \\ 
& & Best dropout& $34.02 \pm 1.48$ & $69.81 \pm 0.56$ \\ 
& & feature grouping & {$\mathbf{68.79 {\pm} 1.04}$} & {$\mathbf{76.45 {\pm} 0.52}$} \\
\cline{2-5}
& {\multirow{1}{*}{CNN} }  & dropout, $p=.5$ & $56.89 \pm 1.62$ & $54.35 \pm 0.93$ \\
\hline
\end{tabular}
\caption{Average and standard error of test accuracy for different regularizers at various noise levels for the Olivetti and HCP-small data sets for logistic regression (LR), MLP and CNN models.}
\label{table:noisy_settings}
\end{table}

\subsection{Results in noisy settings}
In order to explore robustness of classification approaches with different regularizers, we add zero-mean Gaussian noise with varying standard deviations. Here, we use Olivetti faces and HCP-small data sets. The SNR (the ratio of power of signal to noise) of Olivetti and HCP-small becomes $3$ dB and $-2$ dB respectively with moderate additional noise. These values reduce to $0.6$ dB and $-5$ dB  with severe additional noise. We trained three architectures using different regularizers for three noise levels (none, medium and high). We report the test accuracy results with average and standard error computed over $10$ experiments in Table \ref{table:noisy_settings}. We report the best results across $\ell_2$ and dropout parameters.

For the Olivetti faces dataset, MLP with dropout outperforms other
architectures and regularizers when there is no additive Gaussian noise.
However, it does not retain its performance as Gaussian noise is added.
Architectures trained with feature grouping, on the other hand, are
robust to increasing noise level.
Although CNNs do not have an impressive performance, their performance
degrades less quickly with noise compared to the other architectures with $\ell_2$ and dropout.

Similarly, for HCP-small data set, architectures with feature grouping are more robust against additive Gaussian noise. Unlike the Olivetti data set, CNNs perform poorly for all noise settings. This could be because we use 2D convolutions instead of 3D convolutions for a 3D data set. However, 3D convolutions demand much more memory than our available computational resources. Furthermore, the \textit{translation invariance} property of CNNs does not help for brain images, and is in fact detrimental.

\begin{figure}[!t]
\mbox{
\subfigure[Olivetti Faces]{\includegraphics[scale=0.26]{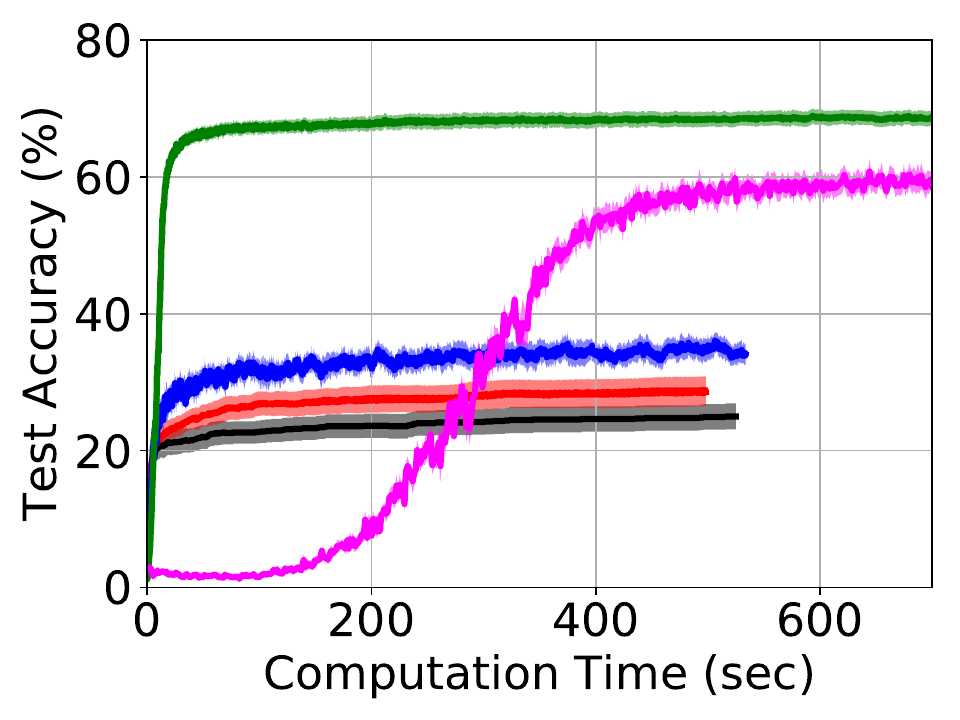}%
    \label{fig:face_mlp_high_noise_acc}}}%
    \mbox{\subfigure[HCP-small]{\includegraphics[scale=0.26]{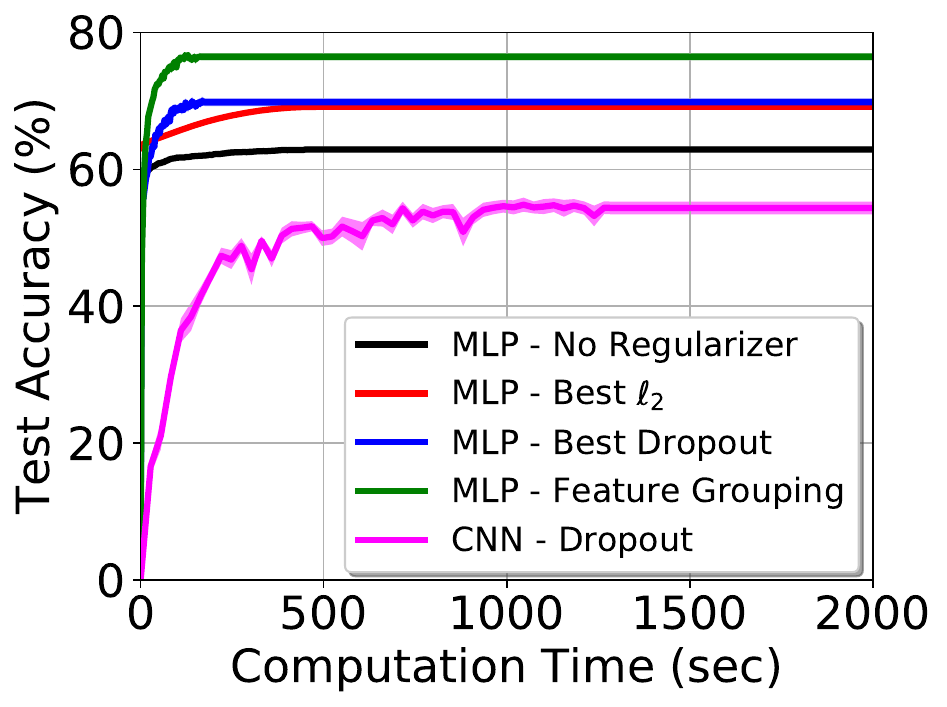} \label{fig:small_hcp_sigma}}} 
\caption{\textbf{Noisy settings:} Performance in terms of test accuracy as a function of computation time for neural networks using feature grouping and best parameters for other regularizers with high noise (a) Olivetti Faces (b) HCP-small.}  
\label{fig:computational_time}
\end{figure}
We  compare the computational performances of CNN with
dropout and MLP with different regularizers for Olivetti faces and HCP-small under
high noise settings. Figure \ref{fig:computational_time} clearly shows
that MLP with feature
grouping achieves higher accuracy in shorter time despite the high
noise for both data sets.

\begin{figure}[!h]
\mbox{
\hspace*{-1ex}%
\subfigure{\includegraphics[scale=0.12, trim={0
1.5cm 0 1.5cm},clip]{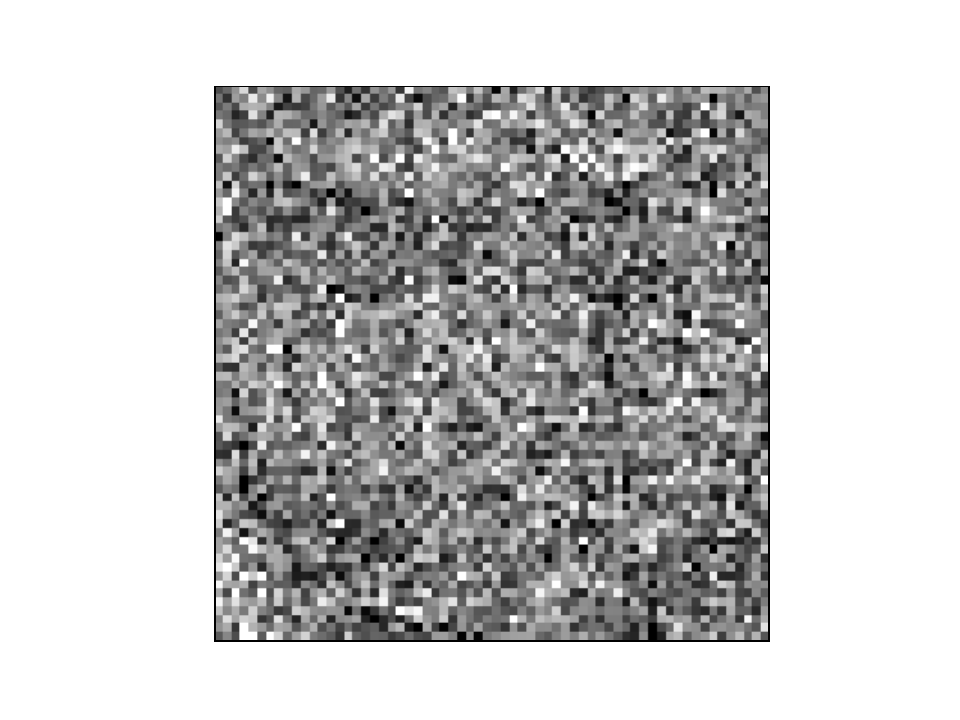} %
    \label{fig:weights_none}} }%
    \hspace*{-1.4ex}%
    \mbox{\subfigure{\includegraphics[scale=0.12, trim={0
1.5cm 0 1.5cm},clip]{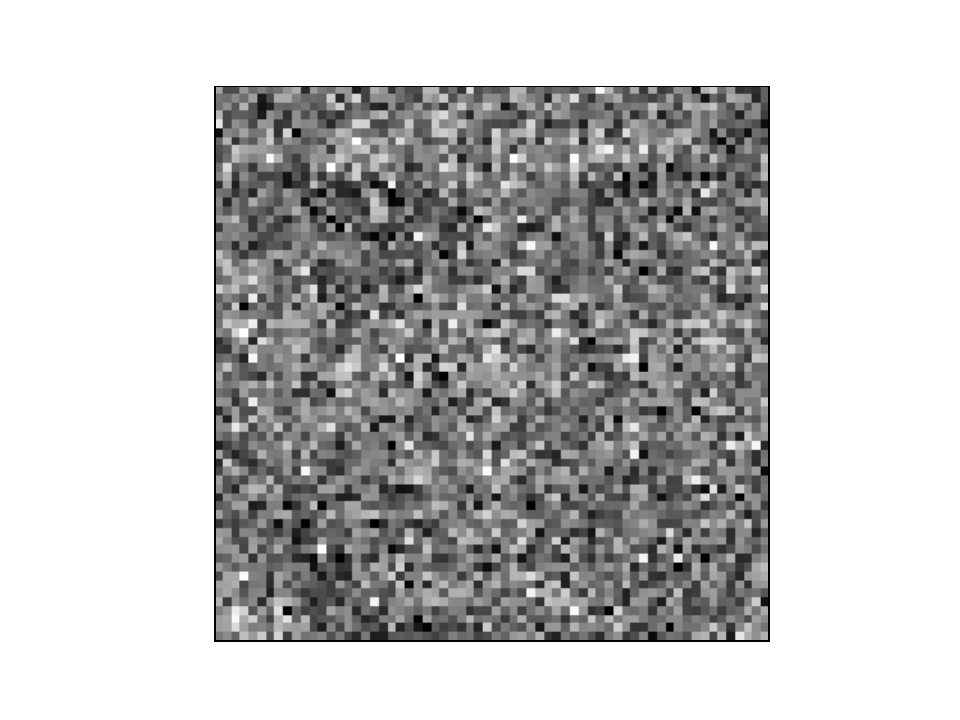} \label{fig:weights_l2}}} 
    \hspace*{-2ex}%
    \mbox{\subfigure{\includegraphics[scale=0.12, trim={0
1.5cm 0 1.5cm},clip]{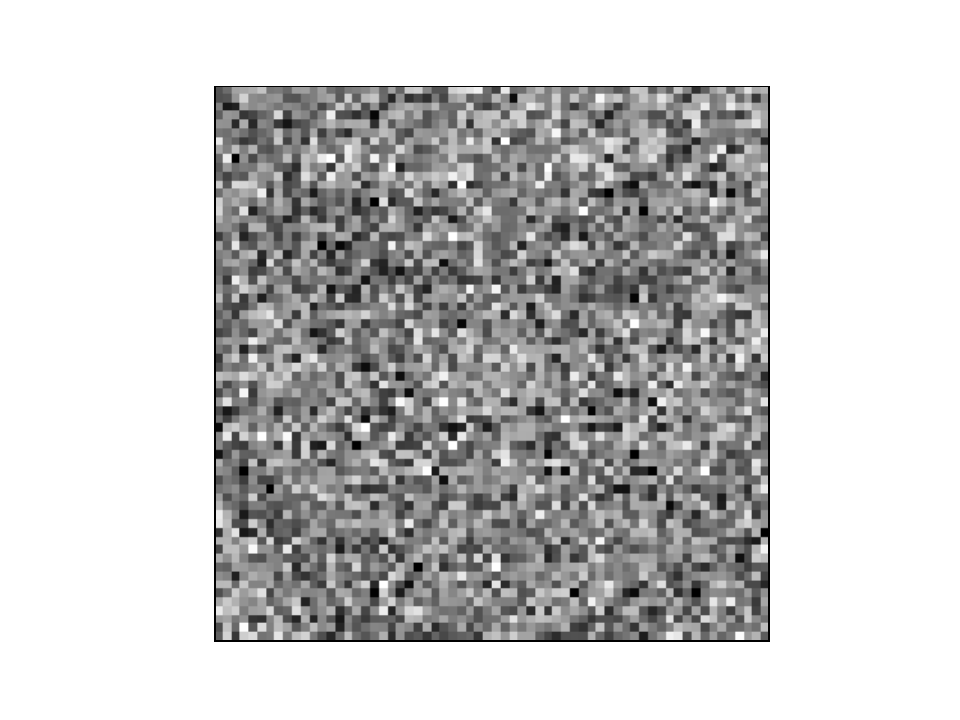} \label{fig:weights_dropout}}}    
    \mbox{\subfigure{\includegraphics[scale=0.12, trim={0
1.5cm 0 1.5cm},clip]{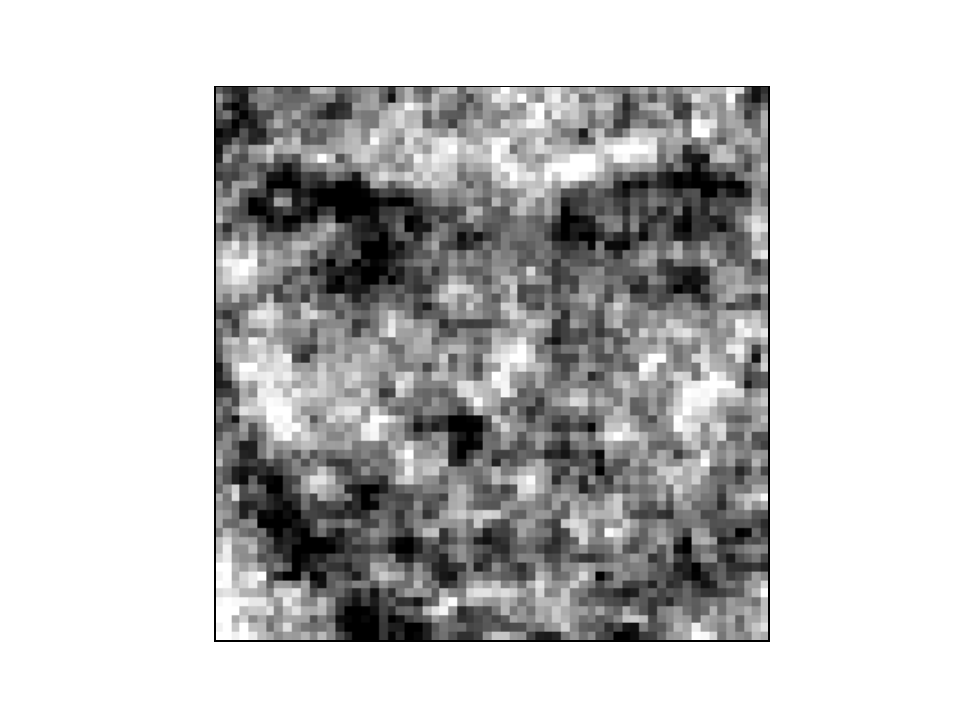} \label{fig:weights_fg}}}%

  \mbox{ 
  \stackunder[1ex]{\subfigure{\includegraphics[scale=0.315]{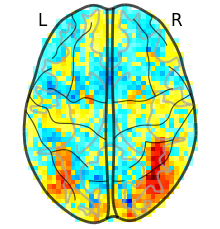}}}{(a)}  \hfill 
   \stackunder[1ex]{\subfigure{\includegraphics[scale=0.315]{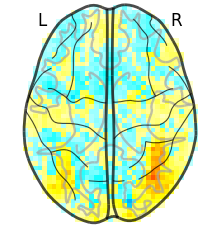}}}{(b)}
    \hfill
    \stackunder[1ex]{\subfigure{\includegraphics[scale=0.315]{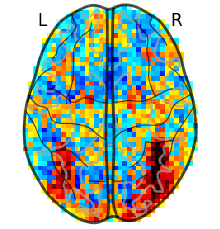}}}{(c)} \hfill
        \stackunder[1ex]{\hspace{2ex} \subfigure{\includegraphics[scale=0.315]{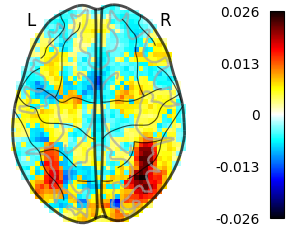}}}{(d)}
}

\caption{Visualization of the learned weights for logistic regression. Top: Olivetti faces data set with high noise level for an arbitrarily selected subject. Bottom: a single subject while performing FACES task. (a) No regularizer (b) Best $\ell_2$ (c) Best Dropout (d) Feature Grouping}
\label{fig:weights}
\end{figure}

We also show in Figure \ref{fig:weights} the learned weights averaged
over $10$ different initializations for each approach. The weights from the
feature grouping approach visually look less noisy, which explains the
superior performance of this approach in noisy settings.
\vspace{-0.6cm}
\subsection{Results for small-sample settings}
\begin{figure}[!h]
\mbox{
\subfigure[Logistic Regression]{\includegraphics[scale=0.26]{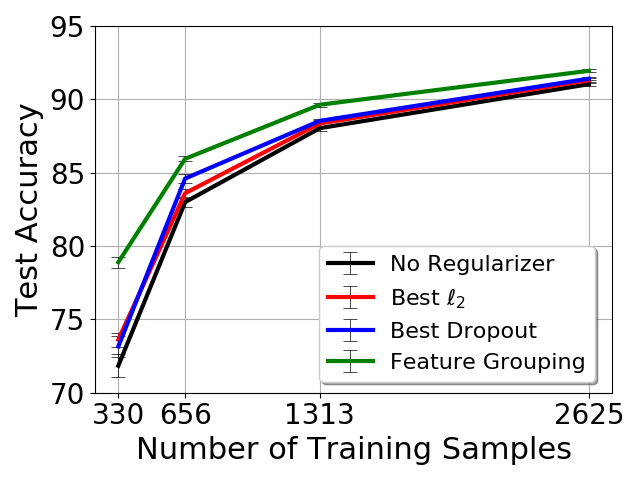}%
    \label{fig:hcp_best_vs_n_samples_lr}}}%
    \mbox{\subfigure[Neural Networks]{\includegraphics[scale=0.26]{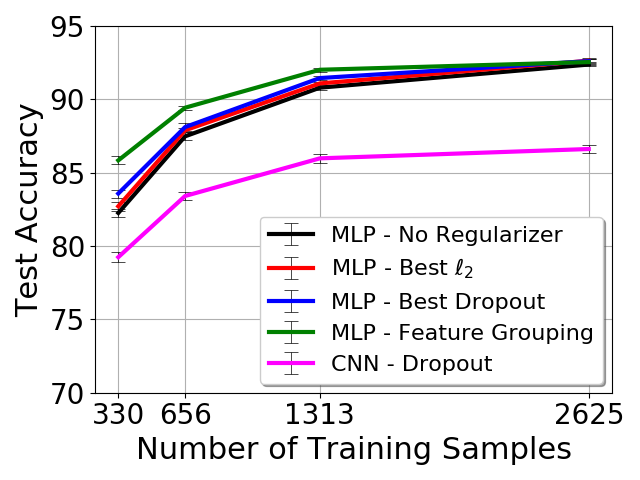} \label{fig:hcp_best_vs_n_samples_with_cnn}}} 
\caption{\textbf{Small-sample settings:} Performance in terms of test
accuracy as a function of number of samples using feature grouping and
best parameters for other regularizers, for HCP data set.}%
\label{fig:hcp_best_vs_n_samples}%
\end{figure}

We explore the robustness of different approaches in small-sample
settings for the HCP data set. Figure \ref{fig:hcp_best_vs_n_samples} shows
logistic regression and MLP learning curves with each regularizer:
the test accuracy when different numbers of samples are used.
It shows that feature grouping clearly outperforms the other approaches both for logistic regression and MLP when  fewer samples are used.
MLPs perform better than logistic regression even for small sample sizes. Similar to the results from HCP-small, CNNs do not perform well on this data set. The difference between feature grouping, $\ell_2$, and dropout disappears as the number of samples increases.

\section{Conclusion}

We propose a new stochastic regularizer, based on feature clustering and
averaging, randomized inside an SGD loop. Our regularizer directly
exploits structure in the data by constructing clusters of correlated
features. This makes it particularly well-suited to data with very high
dimensionality. Unlike classic structured regularizers, our approach can
be plugged into any model, including non-convex ones, solved by gradient
descent. In deep architectures, it operates on the input layer, which is likely to contain more parameters than subsequent layers in high-dimensional problems.

We demonstrate the effectiveness of our regularizer on two problems with
structure in the features: frame-aligned faces and fMRI. In both cases, our method outperforms dropout, $\ell_2$ regularization, and convolutional neural networks when the noise level increases. On the fMRI data, we also show that our method performs best as the sample size decreases.

Our approach comes with little computational cost: it only adds to the
SGD update loop a cost linear in the feature dimension, but reduces the
memory usage of subsequent steps.
Experimental results confirm that neural networks trained 
with our regularizer converge in the same amount of time as with other regularizers, but with higher accuracy.

Regularizations can be seen as modifying the objective function optimized
during training. Our stochastic regularizer forces the model to learn
from smoothed inputs. Its effect decomposes into interpretable
components: the loss on the smoothed inputs, and a regularization term which shrinks model weights for the noisiest feature clusters. We also draw connections to dropout.

The approach proposed here introduces new ideas for developing structured
regularizations, departing from the classic framework of engineering
penalties. Our findings suggest that
the combination of structured random matrices and stochastic optimization
for regularization should be explored further as it is versatile and
computationally efficient. 
The approach should be tested on other data that present a strong stationary structure, such as spectra in chemistry.  

\newpage
\bibliography{mybib}
\bibliographystyle{icml2019}

\newpage
\renewcommand\thefigure{A.\arabic{figure}}  
\renewcommand\thetable{A.\arabic{table}}    
\begin{table*}
\centering
\begin{tabular}{|c|c|c|c|}
\hline
Cognitive Task & Stimuli & Description \\
\hline
\multirow{5}{4em}{Working Memory} & 2BK-0BK& remembering two pictures back versus the current one  \\
& BODY-AVG & presented body parts versus other visual objects \\
& FACE-AVG & presented faces versus other visual objects \\
& PLACE-AVG & presented places versus other visual objects \\
& TOOL-AVG & presented tools versus other visual objects \\
\hline

\multirow{2}{4em}{Gambling} & PUNISH & loss trials when asked to guess a range of a number.  \\
& REWARD & reward trials when asked to guess a range of a number.  \\
\hline

\multirow{5}{4em}{MOTOR } & LF-AVG& Left foot movement versus other movements   \\
& LH-AVG & Left hand movements versus other movements \\
& RF-AVG & Right foot movements versus other movements \\
& RH-AVG & Right hand movements versus other movements  \\
& T-AVG & Tongue movements versus other movements  \\
\hline

\multirow{2}{4em}{Language} & MATH & complete addition and subtraction problems  \\
& STORY & asked questions about topic of the story  \\
\hline

\multirow{2}{4em}{Relational} & MATCH & decide if objects match  \\
& REL & find differences between objects   \\
\hline

\multirow{2}{4em}{Emotion} & FACES & decide which two faces match  \\
& SHAPES & decide which two shapes match   \\
\hline

\multirow{2}{4em}{Social} & RANDOM & presented video clips where objects moved randomly  \\
& TOM & presented video clips where objects interacted   \\
\hline
\end{tabular}
\caption{Cognitive tasks and contrasts used for supervised classification for the neuroimaging data set}
\label{table:tasks}
\end{table*}

\begin{table*}
\centering
\begin{tabular}{|c|c|c|c|}
\hline
Cognitive Task & Stimuli & Description \\
\hline
\multirow{1}{8em}{Working Memory} & 2BK-0BK& remembering two pictures back versus the current one  \\
\hline

\multirow{2}{4em}{Gambling} & PUNISH & loss trials when asked to guess a range of a number.  \\
& REWARD & reward trials when asked to guess a range of a number.  \\
\hline

\multirow{2}{4em}{Relational} & MATCH & decide if objects match  \\
& REL & find differences between objects   \\
\hline

\multirow{2}{4em}{Emotion} & FACES & decide which two faces match  \\
& SHAPES & decide which two shapes match   \\
\hline

\multirow{1}{4em}{Social} 
& TOM & presented video clips where objects interacted   \\
\hline
\end{tabular}
\caption{Cognitive tasks and contrasts used for supervised classification for the neuroimaging data set}
\label{table:small_tasks}
\end{table*}

\begin{table*}
\centering
\rowcolors{2}{gray!15}{white}%
\begin{tabular}{|c|c|c|}
\hline
 r : number of samples used for each $\boldsymbol{\Phi}$ & b: number of projection matrices & Accuracy \\
\hline
10 & 100 & $79.88 \pm 0.53$ \\
10 & 500 & $78.28 \pm 0.54$ \\
50 & 100 & $79.24 \pm 0.66$ \\
50 & 500 & $78.05 \pm 1.04$ \\
50 & 1500 & $78.82 \pm 0.79$ \\
100 & 100 & $78.84 \pm 0.82$ \\
100 & 500 & $77.97 \pm  0.823$ \\
100 & 1500 & $78.50 \pm 0.68$ \\
200 & 100 & $78.30 \pm 0.19$ \\
200 & 500 & $78.46 \pm 0.30$ \\
200 & 1500 & $78.66 \pm 0.60$ \\
\hline
\end{tabular}
\caption{Average and standard deviation of accuracy results of feature grouping for HCP-small dataset with for different $r$ and $b$ values using logistic regression. Number of samples used is 330.}
\label{table:rb_accuracy}
\end{table*}

\end{document}